# BELIEF IN BELIEF FUNCTIONS:
## AN EXAMINATION OF SHAFER'S CANONICAL EXAMPLES

Kathryn Blackmond Laskey
Decision Science Consortium, Inc.
7700 Leesburg Pike, Suite 421
Falls Church, VA 22043[1]

### Abstract

In the canonical examples underlying Shafer-Dempster theory, beliefs over the hypotheses of interest are derived from a probability model for a set of auxiliary hypotheses. Beliefs are derived via a compatibility relation connecting the auxiliary hypotheses to subsets of the primary hypotheses. A belief function differs from a Bayesian probability model in that one does not condition on those parts of the evidence for which no probabilities are specified. The significance of this difference in conditioning assumptions is illustrated with two examples giving rise to identical belief functions but different Bayesian probability distributions.

1. **Introduction**

The artificial intelligence community is in the midst of a lively debate over the representation and manipulation of uncertainty. There is a growing recognition of the need for a numerical uncertainty calculus for at least some classes of problems. A glance through last year's proceedings for this workshop identifies two leading contenders for this role: probability theory and the Shafer-Dempster theory of belief functions. These two theories are closely related. Belief functions are based on probabilities, although common interpretations of beliefs as lower bounds on probabilities are easily shown to be incorrect.

The theory of belief functions lacks the axiomatic foundation of probability theory, but (not coincidentally) Shafer argues that such axiomatic justifications are unnecessary. People do not come to a problem with beliefs "in their heads," waiting to be elicited. Rather, argues Shafer, beliefs should be constructed in a process of comparing one's real-life problem to canonical examples of a theory. Bayesian theory, says Shafer, is based on canonical examples in which the truth is generated according to known chances. In contrast, belief functions arise from canonical examples in which the *meaning of the evidence* is governed by chance. Choose the theory, says Shafer, for which the canonical examples best match your problem.

This constructive approach has obvious appeal, and has been advanced by Bayesian thinkers as well (e.g. Pratt, Raiffa and Schlaiffer, 1964). With this approach, Bayesian theory is dethroned from a lofty (but undeserved) position: there is no uniquely justifiable inference theory. Bayesian theory applies only to the extent that the problem at hand "resembles" the theory's canonical examples.

--------------------

1. Work supported in part by U.S. Army Communications-Electronics Command, Contract No. DAAB07-86-C-A052.

Yet the soundness of a constructive argument depends on more than whether a problem matches a theory's examples. We also need a compelling rationale for why the theory operates as it does on its canonical examples. In this context, the role of axiomatic derivations becomes clear: they form a compelling link between Bayesian theory and certain canonical examples. Is there an equally compelling connection between Shafer's canonical examples and the operations his theory performs on the examples? Shafer appeals to intuition--he argues that the beliefs his theory assigns seem reasonable given the statement of the canonical examples. Of course, some residue of any theory's justification must be left to our intuitions about what are "reasonable" properties. Yet Shafer's examples are complex, and the intuitive leap from example statement to beliefs may not be immediately obvious.

This paper attempts to clarify the connection between examples and theory by comparing Shafer's model of his examples to a Bayesian approach to the same examples. We shall see that belief function models are *incompletely specified* probability models, in which probabilities are specified for only some aspects of a problem (Shafer, 1984). Belief function models are incoherent from the Bayesian perspective because we are not allowed to condition on that part of the evidence which is not modeled probabilistically. On the other hand, coherence would require a willingness (at least in principle) to assign probabilities to any event, whether or not it makes sense to do so. The difference between the two approaches is made concrete by illustrating two situations which could never lead to the same probability model but which give rise to identical belief functions.

## 2. A Canonical Example

The story of the coded message is used repeatedly by Shafer as a canonical example. We begin with two (finite) sets, S and T, called "frames of discernment" by Shafer. The set T is the set of hypotheses of interest to us. We may think of the "truth" $t \in T$ as only partially observable, in the sense that someone other than ourselves has ascertained that t lies in some subset $A \subseteq T$. We think of the subset A as a plaintext message, "the truth lies in A." The set S contains codes, or functions which carry plaintext messages A to some set Q (i.e., an element s of S is a function $s: 2^T \to Q$). We do not get to observe A. Instead, a code s is chosen randomly according to a probability distribution $P(\cdot)$, and we receive the message $q = s(A)$. For concreteness, let us consider

> *Example 1*: A spy is sent to discover whether the enemy intends to attack at dawn. The spy observes a non-empty subset of T = {yes, no} (i.e. she may observe A = {yes}, the enemy will attack; A = {no}, the enemy will not attack; or A = T, she is unable to determine whether the enemy will attack). The set of messages is Q = {APPLE, BANANA, CHERRY}. The possible codes are $s_1$ and $s_2$, with probabilities $P(s_1) = 1/3$, and $P(s_2) = 2/3$, and values:
>
> $s_1(\{yes\})$ = APPLE; $s_1(\{no\})$ = CHERRY; $s_1(T)$ = BANANA; and
> $s_2(\{yes\})$ = APPLE; $s_2(\{no\})$ = BANANA; $s_2(T)$ = CHERRY.
>
> The spy sends the message q = BANANA. The problem is to determine a reasonable degree of belief for whether the enemy will attack.

Shafer's other canonical examples have a similar structure. The cover stories differ, but in each case there are two frames, S and T. Prob-

40

abilities $P(\cdot)$ are defined on S, but not on the hypotheses of interest T. The evidence consists of the probability distribution $P(\cdot)$ and a subset E of $S \times 2^T$ which acts as a constraining relation: if we learned which $s \in S$ was chosen, we would know that t belonged to some subset $A \subseteq T$ for which $(s,A) \in E$. In Example 1, $E = \{(s_1,T), (s_2,\{no\})\}$.

A belief function on T is derived by extending the probability distribution $P(\cdot)$ to the power set $2^T$. Define $S_1 = \{s: (s,A) \in E$ for some $A\}$ to be the set of those s which could have occurred given the evidence E. Define $C(s) = \cup\{A: (s,A) \in E\}$ to be the set of those t compatible with s given E. A basic probability distribution is defined over $2^T$ by

$$m(A) = \sum_{C(s)=A} P(s|S_1) . \tag{1}$$

The belief function is computed from the basic probability distribution by

$$Bel(A) = \sum_{B \subset A} m(B) . \tag{2}$$

In Example 1, $C(s_1) = T$ and $C(s_2) = \{no\}$. All codes are possible, so $S_1 = S$, and the basic probabilities are

$$m(\emptyset) = m(\{yes\}) = 0; \quad m(\{no\}) = 2/3; \quad m(T) = 1/3. \tag{3}$$

Belief in an attack is $Bel(\{yes\}) = 0$; belief in no attack is $Bel(\{no\}) = 2/3$. To the extent that beliefs fail to sum to 1, belief is not committed directly to singleton hypotheses (in this example, belief of 1/3 is committed to T, but not allocated between its elements).

Thus, in the coded message example, the basic probability of a set A is determined by summing the probabilities for all those codes that decode to the set A, after first conditioning on the set of codes that are possible given the message q (this step ensures that basic probabilities sum to 1). Shafer regards it as intuitively clear that (2) defines reasonable degrees of belief for the coded message example. Bel(A) is "interpreted as the probability that the evidence means or implies" that $t \in A$ (in press, p. 31), and "m(A) is, in a certain sense, the total chance that the true message was A," (1981, p. 5). This does not mean, he adds in a note, that A is drawn randomly from the distribution $m(\cdot)$: "It is just that m(A) is the sum of the chances for those codes that decode our encoded message to A." Despite this caveat, readers are likely to interpret m(A) as the (Bayesian) probability that the true message was A. We shall see later that this interpretation is wrong.

3. <u>Dempster's Rule</u>

The basic combination rule for belief functions has a natural interpretation within the formalism of Shafer's canonical examples. Suppose S and U are two sets of codes, and codes $s \in S$ and $u \in U$ are chosen independently according to distributions $P(\cdot)$ and $R(\cdot)$, respectively. We observe coded messages $q_1 = s(A_1)$ and $q_2 = u(A_2)$, where $A_1$ and $A_2$ are the contents of the true plaintext messages. We now have two constraining relations, $E_1 = \{(s,A_1): s(A_1)=q_1\}$ and $E_2 = \{(u,A_2): u(A_2)=q_2\}$. We may combine these into a single set E as follows. Suppose $(s,A_1)$ and $(u,A_2)$ are the true codes and plaintext messages. This means that the truth t is constrained to lie in both $A_1$ and $A_2$, that is, in their intersection

41

$A = A_1 \cap A_2$. Ruling out impossible combinations, we obtain a combined constraining relation $E = \{(s,u,A): A \neq \emptyset, A = A_1 \cap A_2, s(A_1) = q_1, u(A_2) = q_2\}$. The probability distribution over $S \times U$ is given by the product distribution $\Pr(s,u) = P(s) \cdot R(u)$ (because the codes were chosen independently), and extended to a belief function exactly as before. The set $SU_1$ is defined as $\{(s,u):(s,u,A) \in E \text{ for some } A\}$, the set of codes compatible with $E$. The set $C(s,u)$ is defined as $\cup\{A: (s,u,A) \in E\}$, the set of $t$ compatible with codes $s$ and $u$. Then

$$m(A) = \sum_{C(s,u)=A} \Pr(s,u|SU_1) = \frac{1}{\Pr(SU_1)} \sum_{C(s,u)=A} P(s)R(u) \ . \qquad (4)$$

Conditioning on the set $SU_1$ of compatible codes is the usual normalization step of Dempster's Rule. The belief function then comes from (2).

## 4. Belief Functions and Bayesian Probabilities

Equation (4) makes it clear that both the construction of belief functions from Shafer's canonical examples and the application of Dempster's Rule follow from the framework outlined in Section 2. The basic elements are (i) two frames $S$ [$S \times U$ in (4)] and $T$; and (ii) evidence consisting of a probability distribution on $S$ and a constraining relation $E$ which specifies which subsets of $T$ are compatible with each $s \in S$. No probabilities are defined over $T$. To derive degrees of belief over $T$, we first form a conditional probability distribution over $S$, ruling out elements incompatible with the evidence $E$. We then compute the degree of belief in $A$ by summing the conditional probabilities for $s$ such that $A$ contains the set $C(s)$ of $t$ compatible with $s$.

Shafer (in press) argues that this extension of the distribution $P(\cdot)$ to $\text{Bel}(\cdot)$ is entirely natural when $S$ "mediates the interaction" between the evidence and $T$. This means, he says, that "the evidence bears on the question considered by $T$ only indirectly, through its relevance to the question considered by $S$." That this is the case in the coded message example is regarded by Shafer as obvious.

To a Bayesian, this is a curious argument: the set $E$ gives information *jointly* about $S$ and $T$. Suppose, prior to observing the message $q$, we had placed a distribution over the possible plaintext messages $A$. Assume the codes are chosen independently of the plaintext message $A$ (a clear implication of the sense of the example). If the codes are one-to-one (i.e. a single $A$ compatible with each $s$ given the message $q$), then it can be shown that $m(A)$ is the posterior probability of $A$ given $E$, *if all $A$ were a priori equally likely* (Williams, 1982). Otherwise, the interpretation of $m(A)$ as the posterior probability of $A$ is erroneous. Because $E$ contains information about both the code $s$ and the plaintext message $A$, the posterior probabilities depend on *both* the probabilities $P(\cdot)$ over $S$ and the prior probabilities of subsets $A \subset T$.

The result cited above means that a belief function model for the coded message example with one-to-one codes is *mathematically equivalent* to a Bayesian model in which plaintext messages $A$ are assigned equal prior probabilities, and $m(A)$ is the posterior probability of $A$ given $E$. Despite this formal equivalence, Shafer argues that belief function models are *not* Bayesian--"they have their own logic." The Bayesian argument applies only when we think it appropriate to assign probabilities to the plaintext messages. If, on the other hand, it seems appropriate to attach probabilities to the codes but not to the plaintext messages or to t∈T, then we should use the probabilities over S to define beliefs for T as in (1) and (2). We condition only on those aspects of the evidence for which we have specified a probability model.

To a Bayesian, it would always be possible *in principle* to put probabilities on the plaintext messages. If these probabilities are not specified, the Bayesian would either assign them by assumption (e.g., Laskey et al., 1986) or produce an analysis of how the posterior probabilities would depend on the priors if they were specified. Let us try such an analysis for Example 1. The posterior probability of {yes} is always 0; if the prior probabilities of {no} and T are in ratio a:1, their posterior probabilities are in ratio

$$\frac{Pr(A=\{no\}|E)}{Pr(A=T|E)} = \frac{Pr(A=\{no\};E)}{Pr(A=T;E)} = \frac{Pr(s_2,\{no\})}{Pr(s_1,T)}$$

$$= \frac{2a}{1} . \qquad (5)$$

A ratio of 2:1, as in the Shaferian belief function, means to a Bayesian an *implicit* assumption of a=1, or equal prior probabilities.

On reflection, Equation (5), a special case of Williams' result, should not seem surprising. A Bayesian model of the coded message problem requires a complete probability model over $S \times 2^T$. Our model was based on two assumptions, both of which seem to arise naturally from the sense of the example. The first is that the codes are chosen independently of the plaintext message (the spy's actual observation). The second assumption is that the codes are chosen at random with *prior* probability (i.e. before conditioning on the content of the coded message) $P(\cdot)$. For the Bayesian posterior probability to match the belief function, we would need a third requirement: that the posterior probability of each plaintext message given the observed message BANANA is equal to the prior probability of the code transforming the plaintext message into the message BANANA. It is not surprising that this third condition holds only when the prior odds ratio is equal to 1.

But Shafer clearly does not intend the basic probability of a set to be the posterior probability given E that it was the true plaintext message. Rather, as discussed above, a belief function is an incompletely specified probability model. Probabilities are specified for the codes, but no probability model is given for T, either unconditionally or conditional on the codes in S. Since there is no joint distribution over $S \times 2^T$, we may not condition on the full evidence E. Rather, we condition the distribution over S on the set $S_1$ of codes which *could* have occurred given the message q. Shafer defends this as an intuitively plausible way to assign beliefs when no probability model is given for T or its relation to S.

We see, then, that a Bayesian and a Shaferian would operate differently on Shafer's own canonical example. Given a partially specified probability model, the Shaferian adopts a partial conditioning approach that avoids the issue of probabilities on plaintext messages. The Bayesian regards the Shaferian's results as *implicitly* determining probabilities on the plaintext messages. Dempster's Rule follows from Shaferian partial conditioning; to a Bayesian, it requires implicit but untenable assumptions (equal prior probabilities on the component plaintext

43

messages; and that learning one plaintext message $A_1$ tells us nothing
about the other plaintext message $A_2$).

## 5. A Second Example

We have established that belief functions are not Bayesian. Our second
example shows why it matters: the example results in the identical
belief function as Example 1, but the Bayesian posterior probabilities
must be different.

> *Example 2*: The same spy as in Example 1 observes the same plain-
> text message, and again sends a message using one of two possible
> codes. The first code of Example 1 is replaced by a new code that
> distinguishes only between confirmation and nonconfirmation of the
> attack; the second code is left unchanged. The codes are:
>
> $s_1'(\{yes\})$ = APPLE;  $s_1'(\{no\})$ = $s_1'(T)$ = BANANA.
> $s_2(\{yes\})$ = APPLE;  $s_2(\{no\})$ = BANANA;  $s_2(T)$ = CHERRY.

The probabilities of the codes are the same as in Example 1: $P(s_1)$ =
1/3 and $P(s_2)$ = 2/3.

In this example, the constraining relation is given by $E'$ = {$(s_1',\{no\})$,
$(s_1',T)$, $(s_2,\{no\})$}. The compatibility sets, however, are the same as
in Example 1: $C(s_1')$ = T and $C(s_2)$ = {no}. Thus, the basic probability
assignments (3) apply to this example, and belief in no attack is again
2/3.

A Bayesian analysis of this example begins with the same prior odds
ratio a:1 for {no} and T as for Example 1. The posterior odds ratio is
then

$$\frac{Pr(A=\{no\}|E)}{Pr(A=T|E)} = \frac{Pr(s_1',\{no\}) + Pr(s_2,\{no\})}{Pr(s_1',T)} \qquad (6)$$

$$= \frac{a(1/3) + a(2/3)}{1/3} = \frac{3a}{1} \; .$$

In other words, in Example 1, learning that the message was BANANA mul-
tiplies by a factor of 2 the prior odds ratio for plaintext message {no}
versus T. In Example 2, the prior odds ratio is multiplied by 3.
Regardless of prior probability, no Bayesian would assign equal beliefs
in these two examples, yet the resulting belief functions are identical.

Example 2 differs from the examples used by Shafer in that there is no
unique decoding of the code $s_1'$. But to disallow the example by
Shafer's criterion, we would have to argue that S' does not mediate the
interaction between the coded message and T. Yet in Example 1 it is
supposed to be clear that the observed message does bear on T only
through its relevance to which code in S ws chosen. This argument would
not seem to be affected by the double assignment of the codeword BANANA
in Example 2.

## 6. Discussion

To a Bayesian, it is clear that belief in no attack should be higher in the second example than in the first. Everything about the two examples is identical, except that in Example 2, the first code *might* have been used on a plaintext message (no). That is, the spy's having observed that there would be no attack is compatible with either code in Example 2, but only with $s_2$ in Example 1. Therefore, belief in no attack should be higher in Example 2.

A Shaferian would argue differently. In both examples, the second code *implies* that the enemy will not attack. In neither example does the first code imply anything about the enemy's intention (while the spy could have observed (no) in Example 2, she could also have observed T). Therefore, our belief in no attack should be equal to the probability of the second code, and our uncommitted belief should be equal to the probability of the first code.

Which analysis is "correct?" The answer to this question goes to the heart of the philosophical debate between Bayesians and Shaferians. To a Bayesian, even when a probability model is left unspecified, implications of such a model *if it were specified* should be preserved. To a Shaferian, if it makes no sense to specify a probability model for part of a problem, there is no reason to be bound by the implications of such a model.

If you accept equal beliefs for Examples 1 and 2, then you probably accept (1) and (2) as reasonable formulae for deriving beliefs on a frame T from probabilities on a related frame S. Since Dempster's Rule follows from (1) and (2), you should accept that also. If, on the other hand, you feel that belief in no attack should be higher for Example 2, then you disagree with how beliefs are assigned for the canonical examples Shafer uses to justify his theory. In complex problems, this foundational disagreement may have implications that are not so clear as in the simple example of this paper.